%% file: main.tex
\documentclass{article}

\PassOptionsToPackage{numbers, compress}{natbib}

\usepackage[preprint]{neurips_2023}

\usepackage[utf8]{inputenc} 
\usepackage[T1]{fontenc}    
\usepackage{hyperref}       
\usepackage{url}            
\usepackage{booktabs}       
\usepackage{amsfonts}       
\usepackage{nicefrac}       
\usepackage{microtype}      
\usepackage{xcolor}         
\usepackage{graphicx}
\usepackage{amsmath,amssymb} 
\usepackage{multirow}
\usepackage{caption}
\usepackage{ulem}
\usepackage{cleveref}

\title{Fine-grained Text-Video Retrieval with Frozen Image Encoders}

\author{%
Zuozhuo Dai, Fangtao Shao, Qingkun Su, Zilong Dong, Siyu Zhu \\
Alibaba Group \\
\texttt{\{zuozhuo.dzz, shaofangtao.sft, qingkun.sqk,} \\
\texttt{list.dzl, siting.zsy\}@alibaba-inc.com} \\
}

\begin{document}

\maketitle

\begin{abstract}

State-of-the-art text-video retrieval (TVR) methods typically utilize CLIP and cosine similarity for efficient retrieval.
Meanwhile, cross attention methods, which employ a transformer decoder to compute attention between each text query and all frames in a video, offer a more comprehensive interaction between text and videos. 
However, these methods lack important fine-grained spatial information as they directly compute attention between text and video-level tokens. 
To address this issue, we propose CrossTVR, a two-stage text-video retrieval architecture.
In the first stage, we leverage existing TVR methods with cosine similarity network for efficient text/video candidate selection. 
In the second stage, we propose a novel decoupled video text cross attention module to capture fine-grained multimodal information in spatial and temporal dimensions. 
Additionally, we employ the frozen CLIP model strategy in fine-grained retrieval, enabling scalability to larger pre-trained vision models like ViT-G, resulting in improved retrieval performance. 
Experiments on text video retrieval datasets demonstrate the effectiveness and scalability of our proposed CrossTVR compared to state-of-the-art approaches.


\end{abstract}

\section{Introduction}\vspace{-2mm}
Text Video Retrieval (TVR) is a key research task in the field of visual-language understanding. 
Its goal is to retrieve relevant videos in response to a given natural language text and vice versa. 
Generally, there are three main approaches to TVR.
The first approach~\cite{luo2021clip4clip, liu2022ts2,  Wang2022DisentangledRL, fang2021clip2video,ma2022x} involves independent mapping of text and videos to a joint embedding space with the duel encoders, followed by a light-weight cosine similarity calculator for the approximate nearest neighbor search.
While this approach is efficient, its accuracy is limited by the straightforward dot product-based vision-text interaction within the shared embedding space. 
The second approach~\cite{gorti2022x,luo2021clip4clip} uses cross attention transformers to compare each word to the frames in the video, enabling fine-grained interaction between text and videos. 
However, this approach is computationally expensive. 
The third approach, as proposed by Miech et al.~\cite{Miech2021ThinkingFA, Li2021AlignAP,Wang2022OmniVLOF} combines both of the types mentioned above. It involves first eliminating videos that do not have anything in common with the text description using a similarity-based approach. Then, it uses cross attention to obtain the final promising candidates.
Especially, state-of-the-art performance has been achieved using a similarity-based approach combined with multimodal contrastive learning model CLIP~\cite{radford2021learning}. 
In this paper, we investigate the third approach based on the CLIP paradigm~\cite{luo2021clip4clip, liu2022ts2, Wang2022DisentangledRL, gorti2022x, fang2021clip2video,ma2022x}, and propose an efficient fine-grained text-video retrieval method named CrossTVR.


\begin{figure}
    \centering
    \vspace{-2mm}
    \includegraphics[width=0.9\textwidth]{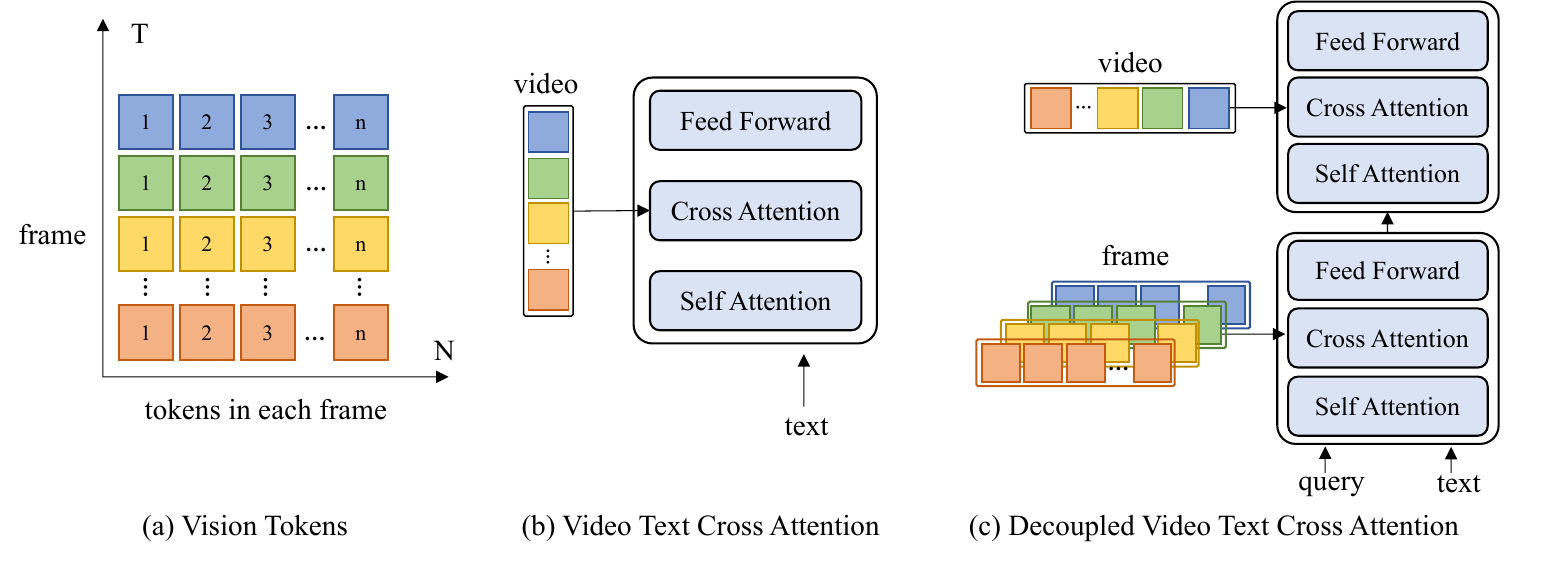}
    \vspace{-2mm}
    \caption{(a) shows given a video with $T$ frames, the vision encoder generates $T\times N$ tokens, where each frame contains $N$ tokens. Tokens with the same color belong to the same frame. Current cross attention based methods (b)~\cite{gorti2022x,luo2021clip4clip,Miech2021ThinkingFA} only compute video level attention. Our method (c) decouples spatial and temporal attention, which computes frame level spatial attention and video level temporal attention separately.}\label{fig:cross}
    \vspace{-6mm}
\end{figure}
  
In Figure ~\ref{fig:cross}, we illustrate the procedure of generating $N$ visual tokens for each of the $T$ frames of a given video, resulting in $T \times N$ vision tokens in total. 
However, directly concatenating all the visual tokens of each frame for cross attention can be quite costly in terms of memory and time. 
As a result, current cross attention based methods~\cite{gorti2022x,luo2021clip4clip,Miech2021ThinkingFA} compute cross attention between text and video tokens, which are generated by simply mean pooling or selecting only the [CLS] token in each frame. 
Unfortunately, this approach can result in the loss of fine-grained spatial information, such as subtle movement and small objects.
To address this problem, we propose a decoupled video text cross attention module that computes frame-level spatial text attention and video-level temporal text attention separately. 
Specifically, the visual tokens generated from the pre-trained image vision encoder for each frame are well-suited for spatial modeling. 
We can respectively compute spatial frame-text cross attention between text tokens and $N$ visual tokens of each frame to extract spatial-enhanced multimodal features. Furthermore, to capture the temporal video-text interaction, we compute cross attention between the multimodal features and representative video tokens (chosen by a token selector) to capture the salient temporal semantic features.

Thanks to recent advancements in Contrastive Language-Image Pre-training (CLIP)~\cite{radford2021learning}, which has been pre-trained on large-scale open-vocabulary image-text pairs, a frozen vision encoder~\cite{Lin2022FrozenCM} and a learnable lightweight transformer decoder can now be used to learn high-quality video representations without relying on an end-to-end fine-tuning regime.
In our proposed decoupled video text cross attention module, a sequence of learnable query tokens is utilized to extract relevant visual features from the vision-language pre-trained model.
This allows us to freeze the vision encoder and only train the cross attention module, resulting in a significant reduction in computation and memory resources, which in turn enables us to adopt very large image architectures, such as the largest open-sourced ViT-G~\cite{EVA-CLIP} vision encoder. 
Based on the main ideas presented above, we propose a two-stage text-video retrieval architecture called CrossTVR.
In the first stage, we utilize a lightweight cosine similarity network, which is trained using ViT-B encoder (utilized in all our experiments).
For the second stage, we adopt the proposed decoupled video text cross attention module that fuses the fine-grained visual and text tokens. 
Combined with hard mining strategy, this module can capture the minor difference of the well-selected candidates and leads to better re-ranking results.

We extensively test our proposed CrossTVR method on multiple text-video retrieval benchmarks, such as MSRVTT~\cite{xu2016msr}, VATEX~\cite{wang2019vatex}, LSMDC~\cite{rohrbach2017movie}, MSVD~\cite{chen2011collecting}, and DiDeMo~\cite{anne2017localizing}. 
Our approach achieves state-of-the-art performance on all of these benchmarks. 
We also demonstrate that our decoupled video text cross attention is compatible with general cosine-similarity methods~\cite{luo2021clip4clip,liu2022ts2,gorti2022x}, which further improve their accuracy by 2.5\% on CLIP4Clip~\cite{luo2021clip4clip}, 2.6\% on TS2Net~\cite{liu2022ts2}, and 1.2\% on X-Pool~\cite{gorti2022x} on Text2Video R@1.
It is worth noting that due to the decoupled cross attention module and frozen CLIP model strategy~\cite{bain2021frozen}, our proposed CrossTVR can be scaled from small (ViT-B) to very large pre-trained models (ViT-G) at a low cost, with an obvious increase 4.4\% in accuracy and a relatively small increase (+17\%) in training time.
This illustrates the superior scalability of CrossTVR, compared to similarity-based approaches that require end-to-end finetuning, along with the advancements of even larger scale pre-trained models in the future.

\input{related}
\input{method}

\input{exp}

\input{conclusion}
\bibliographystyle{splncs04}
\bibliography{egbib}
\input{appendix}
\end{document}

%% file: related.tex
\vspace{-3mm}\section{Related Work}\vspace{-2mm}
\paragraph{Text-Video Retrieval}
Various approaches have been proposed to deal with text-video retrieval tasks,
which generally belongs to one of three types: cosine similarity based, cross attention based, and the combination of both.
Cosine similarity based approaches~\cite{lei2021less,bain2021frozen,luo2021clip4clip,fang2021clip2video,cheng2021improving} usually use contrastive learning from CLIP~\cite{radford2021learning} to compute the cosine similarity between video and text.
CLIP4Clip~\cite{luo2021clip4clip}, CLIP2Video~\cite{fang2021clip2video} and TS2Net~\cite{liu2022ts2} study various mechanism to transfer knowledge from pre-trained CLIP~\cite{radford2021learning} to video retrieval task.
However, the accuracy of cosine similarity based methods is limited due to the simplicity of vision-text interaction model defined by the dot product in the joint embedding space.
Cross attention based methods use attention modules instead, to achieve fine-grained multimodal interaction.
X-Pool~\cite{gorti2022x} proposes a scaled dot product based cross-modal attention model that generates an aggregated video representation conditioned on the text’s attention weights over the frames. 
CLIP4Clip-tightTransf~\cite{luo2021clip4clip} uses cross attention transformer to inference the complex relationship between arbitrary-length text and videos. However, the cross attention transformer suffers from optimization difficulty and high training costs due to computation complexity, which performs no better than cosine similarity based methods.
To combine the advantages of both, some methods~\cite{Miech2021ThinkingFA, Li2021AlignAP,Wang2022OmniVLOF} use a coarse-to-fine strategy, i.e. using fast cosine similarity to get coarse retrieval candidates, and then using cross attention to get the final desired results. We follow the third approach and propose a decoupled spatial temporal cross attention module, which captures both the frame-level and the video-level fine-grained interactions.

\paragraph{Vision-Language Pre-training}
Vision-language pre-trained models have shown promising results in image-language tasks such as image retrieval, visual question answering, and image caption. They usually use a shared self-attention transformer encoder with multi-modal input\cite{li2020unicoder,su2019vl,zhou2020unified}, a cross attention transformer to fuse different modalities~\cite{lei2021less,Li2021AlignBF,Wang2022ImageAA} or contrastive loss to align text and visual embedding~\cite{bain2021frozen,Jia2021ScalingUV,radford2021learning}.
Most recently, FLAVA~\cite{Singh2021FLAVAAF}, BLIP~\cite{Li2022BLIPBL}, and CoCa~\cite{Yu2022CoCaCC} design one vision-language foundation model to support both multi-modal alignment and generation tasks. 
Similarly, video-language pre-trained models~\cite{li2020hero,luo2020univl,Wang2022InternVideoGV,Xue2022CLIPViPAP} are proposed for video-language tasks. ALPRO~\cite{Li2021AlignAP} aims to learn fine-grained alignment between video embeddings and text embeddings. 
Frozen~\cite{bain2021frozen}, OmniVL~\cite{Wang2022OmniVLOF}, and X2-VLM~\cite{Zeng2022X2VLMAP} propose a unified foundation model for image-language and video-language by pre-training on both image and video datasets. These models require additional training on a large amount of pre-training data, which cannot use the off-the-shell vision-language pre-trained models. Following recent works~\cite{luo2021clip4clip, liu2022ts2}, instead of pre-training, our method aims to directly transfer the knowledge from the image-text pre-trained model to video-text retrieval task. Thanks to the frozen CLIP strategy, we can use very large vision-language pre-trained models to further boost up the performance.

\paragraph{Spatial Temporal Attention} 
Spatial temporal attention has widely applied in many video applications, such as video classification~\cite{arnab2021vivit,bertasius2021space}, human pose estimation~\cite{zheng20213d,Zhang_2022_CVPR}, action recognition~\cite{Yang2023AIMAI}, etc.  In video classification, some approaches decouple temporal-spatial attention to extract richer video features. ViViT~\cite{arnab2021vivit} proposes several ways to decouple models along spatial and temporal to increase efficiency and scalability. TimeSformer~\cite{bertasius2021space} also uses a divided attention architecture to apply temporal attention and spatial attention separately in each block of the network.
In video retrieval, many methods directly use spatial temporal transformer to extract video features. TS2Net~\cite{liu2022ts2} and CLIP4Clip~\cite{luo2021clip4clip} use CLIP~\cite{radford2021learning} vision encoder as temporal self attention module to extract spatial-temporal visual features and use cosine similarity to conduct multimodal interaction.
HCMI~\cite{jiang2022tencent} fuses hierarchical video representations with text features by contrastive loss, to achieve multi-level interaction. Further, X-pool~\cite{gorti2022x} uses video text cross attention to get the most similar video frames to the text semantically.
However, the spatial temporal attention model in the video text retrieval task only select [CLS] token or a subset of vision tokens, where important fine-grained features might be lost.
In this paper, we first propose the video text cross attention module, which decouples frame-level spatial and video-level temporal attention for multimodal representation. Our quantitative experiments prove it is more effective than existing spatial-temporal attention methods in TVR task.

%% file: method.tex
\section{Method}\vspace{-2mm}

\begin{figure}
  \centering
  \includegraphics[width=0.9\textwidth]{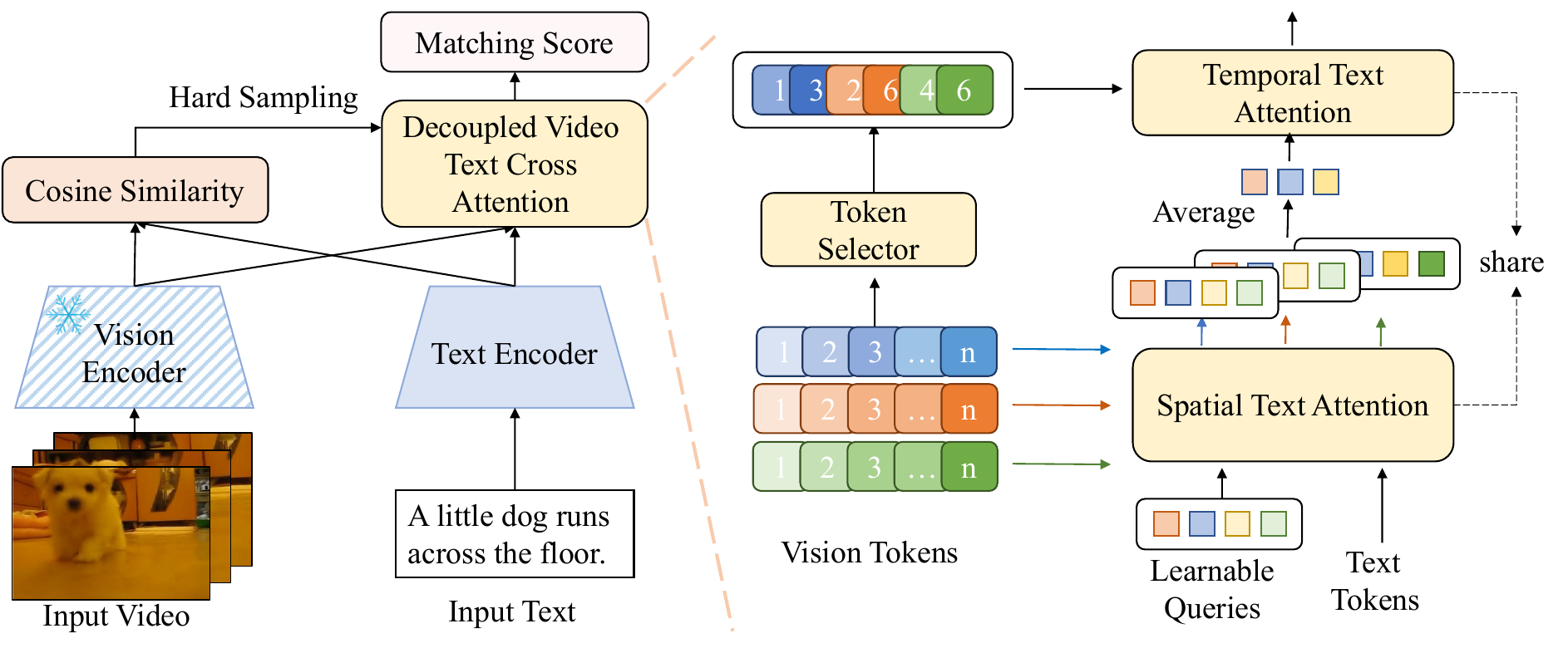}
  \vspace{-2mm}
  \caption{The second stage of CrossTVR training pipeline takes a video with 3 frames as an example. Vision tokens with the same color belong to the same frame. During training, the vision encoder is frozen. We first use the cosine similarity header to select hard negative samples and then compute fine-grained cross attention matching scores between them. The decoupled video text cross attention computes frame-level spatial text attention and video-level temporal text attention separately. Finally, we use a single fully connected layer to output the matching score.}\label{fig:pipeline}
  \vspace{-6mm}
\end{figure}

The objective of text-video retrieval (TVR) is to identify the most relevant videos based on a given text query. 
Our TVR framework comprises two stages of training.
Initially, we train a small-scale cosine similarity-based network (TS2Net~\cite{liu2022ts2} with ViT-B) through contrastive learning for efficient retrieval. 
Subsequently, we train a larger cross attention-based network for more detailed matching.
The second stage of CrossTVR is further elucidated in Figure~\ref{fig:pipeline}, which includes a frozen vision encoder, a text encoder, and a decoupled video text cross attention module.
We utilize a cosine similarity header to generate a similarity matrix for input video-text pairs, enabling us to pick challenging negative pairs for decoupled video text cross attention, as discussed in Section~\ref{sec:spatial_temporal_attention}.
A single fully-connected layer is then employed to generate a matching score. 
The decoupled video text attention module facilitates frame-level and video-level interaction with text, potentially capturing fine-grained multimodal information. 
Additionally, using a frozen vision encoder (Section~\ref{sec:frozen_image_encoder}) makes CrossTVR more scalable to larger pre-trained models. 
The training and inference procedures are detailed in Section~\ref{sec:train} and Section~\ref{sec:inference}, respectively.

\vspace{-3mm}\subsection{Decoupled Video Text Cross Attention}\vspace{-2mm}\label{sec:spatial_temporal_attention}
In this section, we present our proposed architecture for fine-grained video-text cross 
attention that efficiently incorporates both spatial and temporal information.
Existing multimodal Transformer-based TVR methods, such as~\cite{feichtenhofer2019slowfast,luo2021clip4clip}, often compute multimodal attention directly between text and video [CLS] tokens, leading to a loss of frame-level fine-grained visual appearance representations.
To address this issue, we decouple spatial and temporal attention computation by incorporating a spatial-text attention module and a temporal-text attention module, as shown in Figure ~\ref{fig:cross}.
The spatial-text attention module captures the frame-text interaction by computing attention between the given text and each frame, while the temporal-text attention module aims to learn the global video-text multimodal representation by aggregating spatial interaction information of all frames.
This architecture shares the same parameters between the two modules, facilitating parameter-efficient training. 
The decoupled but parameter-sharing design effectively enhances multimodal representation learning and enables spatial and temporal attention to benefit from each other.


\paragraph{Spatial Text Attention} 
We achieve frame-level vision-language fusion using a cross attention module, which extends the formulation of QFormer~\cite{li2023blip2} from a single image to video sequences. 
The module concatenates a fixed number of learnable query embeddings and text embeddings as query input to the cross attention transformer. 
The queries interact through self-attention layers and then with each frame feature of the video through cross attention individually. 
This approach enables queries to retrieve both the overall information (represented by the [CLS] token) and the salient spatial token in each frame. This helps to maintain the fine-grained spatial features of the video as much as possible, thereby facilitating their fusion with textual information.
As a result, we obtain the frame-level spatial enhanced queries for the video using the following equations:
\vspace{-2mm}
\begin{align}\label{eqn:spatial}
    \textbf{z}_{t,1} &= Attn(Concat(Q, X), V_t) \\
    \textbf{z}_{t,l} &= Attn(\textbf{z}_{t,l-1}, V_t),\ l=2,...,L\\
    X_{spatial}(t) &= Avg(\textbf{z}_{t,L}^1, \textbf{z}_{t,L}^2, \cdots, \textbf{z}_{t,L}^{N_Q})
\end{align} 
\vspace{-6mm}

Here, $Q$ represents $N_Q$ learnable query embeddings, $X$ is text tokens, $V_t$ is the vision tokens for the $t_{th}$ frame, $\textbf{z}_{t,l}^i$ is the $i$-th token of the $l$-th spatial cross attention layer's output given the $t$-th frame tokens, and $X_{spatial}(t)$ selects and averages the first $N$ tokens of the last cross attention layer's output $\textbf{z}_{t,L}$. As a result, $X_{spatial}$ of size $T$ is obtained as the spatial enhanced text queries.


\paragraph{Temporal Text Attention} 
We compute the video-level interaction between video and text through temporal-text cross attention, using the spatial enhanced text queries as queries and video feature tokens as values. 
However, the way in which we aggregate multi-frame information has an important impact on video-level interaction. 
Using only [CLS] tokens tends to lose detailed information while introducing all tokens brings irrelevant and redundant information. 
To address this, we draw on the token selector module described in~\cite{liu2022ts2} to obtain the salient objects and movement in terms of temporal. 
As we mentioned before, this module shares parameters with the spatial text attention module for training efficiency. 
Given a sequence of visual tokens, the token selector module employs an MLP followed by a Softmax layer to predict the importance score for each token and select the $M$ most informative tokens for each frame. This can be expressed in the following equations:
  \vspace{-2mm}
\begin{align}\label{eqn:temporal}
  V_{select}(t) &= TokenSelector(V_t, M) \\
  \textbf{z}_1 &= Attn(Concat(X_{spatial}(t)), Concat(V_{select}(t))) \\
  \textbf{z}_l &= Attn(\textbf{z}_{l-1}, Concat(V_{select}(t))),\ l=2,...,L
  \vspace{-6mm}
\end{align} 
Here, the temporal text module computes the cross attention between $M\times T$ flattened visual tokens (as key and values) and $T$ spatial enhanced tokens (as queries) of $\textbf{z}_L$.

\subsection{Frozen Image Encoder}\vspace{-2mm}\label{sec:frozen_image_encoder}
Recently, CLIP~\cite{radford2021learning} has demonstrated promising feature transferability for various downstream image tasks, such as classification~\cite{Zhou2021LearningTP} and segmentation~\cite{Lddecke2021ImageSU}.
To extend CLIP to video tasks, EVL~\cite{Lin2022FrozenCM} and AIM~\cite{Yang2023AIMAI} use frozen CLIP image encoder and a learnable lightweight decoder or a sequence of adaptors to adapt it for single modal video recognition tasks.
In contrast, our cross attention header aims to transfer fine-grained vision features based on text queries for multimodal understanding.
Specifically, in the second stage of CrossTVR, we select two state-of-the-art image pre-trained vision transformers as the frozen vision encoder: ViT-B/32 from CLIP~\cite{radford2021learning} and ViT-G/14 from EVA-CLIP~\cite{EVA-CLIP}. 
We refer to them as CrossTVR Base and CrossTVR Large respectively, for comparison with recent TVR methods~\cite{luo2021clip4clip,liu2022ts2}.
For both CrossTVR Base and CrossTVR Large, since the cross attention header can explore fine-grained relationship between video and text during re-ranking, we can use a relatively smaller vision encoder ViT-B/32 for cosine similarity to ensure efficient retrieval in the first stage. In the second stage of CrossTVR Large, since ViT-G/14 is frozen, the total number of training parameters is significantly smaller than an end-to-end finetuned ViT-G network~\cite{luo2021clip4clip}, reducing GPU memory usage by 91\%. 
Notably, our decoupled video text cross attention and frozen CLIP model strategies complement existing similarity-based approaches~\cite{luo2021clip4clip,liu2022ts2,gorti2022x}.






\subsection{Training}\vspace{-2mm}\label{sec:train}

To train CrossTVR, we follow a two-stage approach. 
In the first stage, we use contrastive learning to train the cosine similarity network, with TS2Net ViT-B/32~\cite{liu2022ts2} being the default choice. 
In the second stage, we freeze the vision encoder and train the cross attention header via video text matching. 
Notably, we utilize separate vision encoders in the two stages of CrossTVR: similarity search and re-ranking.
Since CrossTVR seeks to learn fine-grained interaction between video and text representation, we adopt the hard negative mining strategy from ALBEF~\cite{Li2021AlignBF} to generate informative negative pairs during training.
Here, we consider a negative video-text pair to be hard if they share similar semantics but differ in fine-grained details.
To create informative negative pairs during training, we first compute the similarity matrix for a batch of text-video pairs using the cosine similarity network. 
Then, based on the similarity matrix, we construct a set of positive and hard negative pairs. 
Specifically, for each video in a mini-batch, we sample one negative text from the same batch following the contrastive similarity distribution, where texts with higher matching scores with the video are more likely to be sampled.
Similarly, we also sample one hard negative video for each text. With these training pairs, we compute the text-video matching loss $\mathcal{L}_\mathrm{tvm}$ as follows:
\vspace{-1mm}
\begin{equation}
  \label{eqn:itm}
  \mathcal{L}_\mathrm{tvm} = \mathbb{E}_{(X,V) \sim {D}} \mathrm{H} (\textbf{y}, \textbf{p}(X,V))
\end{equation}
Here, ${D}$ represents the set of positive and negative sample pairs, $\mathrm{H}$ denotes the cross-entropy loss, $\textbf{y}$ is a 2D one-hot vector representing the ground-truth label, and $\textbf{p}$ denotes the decoupled video text cross attention matching score.

\subsection{Inference}\vspace{-2mm}\label{sec:inference}
During the inference process, we begin with computing a rough similarity matrix through the cosine similarity network TS2Net~\cite{liu2022ts2}.
Subsequently, we utilize decoupled video text cross attention to re-rank text-video pairs based on their top similarity scores.
To elaborate further, suppose we have a query text and a vast video database containing $M$ videos. We initially compute $M$ similarity scores between the text and $M$ videos using cosine similarity. 
Then, we select the top $K$ videos (where $K\ll M$) for fine-grained re-ranking using cross attention.
Finally, we determine the final rank of the top $K$ videos by adding the scores of cosine similarity and fine-grained similarity, which is computed using cross attention. 
Since $K$ is considerably smaller than the whole dataset, the inference of TVR can be performed efficiently.
 

%% file: exp.tex
\section{Experiments}\vspace{-2mm}
\subsection{Experimental Settings}\vspace{-2mm}
\textbf{Datasets.} We conduct experiments on following five popular text-video retrieval benchmarks:

\quad $\bullet$ \textbf{MSR-VTT}~\cite{xu2016msr} comprises 10,000 videos, each with 20 captions and a duration of 10 to 32 seconds. 
We follow previous studies~\cite{gabeur2020multi} and adopt the `Training-9K' split, which consists of a training set of 9,000 videos and 180,000 captions, and a test set of 1,000 text-video pairs.

\quad $\bullet$ \textbf{MSVD}~\cite{chen2011collecting} dataset includes 1970 videos and 120,000 captions. 
Each video has around 40 captions and varies in length from 1 to 62 seconds. 
The dataset is divided into three subsets: a training set with 1200 videos, a validation set with 100 videos, and a test set with 670 videos.

\quad $\bullet$ \textbf{VATEX} ~\cite{wang2019vatex} comprises 34,991 video clips, each with multiple captions. Our experiments follow the HGR split used in~\cite{chen2020fine}. 
The training set includes 25,591 videos, while the validation and test set each contain 1,500 videos.

\quad $\bullet$ \textbf{LSMDC}~\cite{rohrbach2017movie} includes 118,081 videos and an equal number of titles obtained from 202 movies, with each video ranging from 2 to 30 seconds in length. The training set contains 109,673 videos, while the validation and test sets include 7,408 and 1,000 videos, respectively. Notably, the test set comprises videos from independent movies that are not included in the training or validation sets.

\quad $\bullet$ \textbf{DiDeMo}~\cite{anne2017localizing} comprises more than 10,000 videos and 40,000 captions, with all videos sourced from Flickr and segmented into 5-second intervals. 
The training, test, and validation sets contain 8,395, 1,065, and 1,004 videos respectively. 
Consistent with prior research~\cite{lei2021less}, all video descriptions are concatenated into a single query for video-paragraph retrieval tasks.

\begin{table}[!t]\small
\centering
\renewcommand{\arraystretch}{1.2}
\setlength{\tabcolsep}{1.2mm}{
\begin{tabular}{c|ccccc|ccccc}
\hline
\multirow{2}{*}{Method} & \multicolumn{5}{c|}{Text2Video} & \multicolumn{5}{c}{Video2Text}  \\ \cline{2-11} 
                        & R@1$\uparrow$  & R@5$\uparrow$   & R@10$\uparrow$  & MdR$\downarrow$  & MnR$\downarrow$   & R@1$\uparrow$   & R@5$\uparrow$   & R@10$\uparrow$  & MdR$\downarrow$  & MnR$\downarrow$   \\ \hline
CE~\cite{liu2019use}                                                & 20.9 & 48.8 & 62.4 & 6.0 & 28.2 & 20.6 & 50.3 & 64.0 & 5.3 & 25.1 \\
MMT~\cite{gabeur2020multi}                                          & 26.6 & 57.1 & 69.6 & 4.0 & 24.0 & 27.0 & 57.5 & 69.7 & 3.7 & 21.3 \\
SUPPORT\cite{patrick2020support}                                    & 27.4 & 56.3 & 67.7 & 3.0 & -    & 26.6 & 55.1 & 67.5 & 3.0 & -    \\
Frozen\cite{bain2021frozen}                                         & 31.0 & 59.5 & 70.5 & 3.0 & -    & -    & -    & -    & -   & -    \\
CLIP4Clip\cite{luo2021clip4clip}                                    & 44.5 & 71.4 & 81.6 & 2.0 & 15.3 & 42.7 & 70.9 & 80.6 & 2.0 & 11.6 \\
CenterCLIP\cite{Zhao2022CenterCLIPTC}                                 & 44.2 & 71.6 & 82.1 & 2.0 & 15.1 & 42.8 & 71.7 & 82.2 & 2.0 & 10.9 \\
CAMoE\cite{cheng2021improving}                                      & 44.6 & 72.6 & 81.8 & 2.0 & 13.3 & 45.1 & 72.4 & 83.1 & 2.0 & 10.0 \\
CLIP2Video\cite{fang2021clip2video}                                 & 45.6 & 72.6 & 81.7 & 2.0 & 14.6 & 43.5 & 72.3 & 82.1 & 2.0 & 10.2 \\
X-Pool\cite{gorti2022x}                                             & 46.9 & 72.8 & 82.2 & 2.0 & 14.3 & -    & -    & -    & -   & -    \\
X-CLIP\cite{ma2022x}                                                & 46.1 & 73.0 & 83.1 & 2.0 & 13.2 & 46.8 & 73.3 & 84.0 & 2.0 & 9.1  \\
TS2-Net\cite{liu2022ts2}                                            & 47.0 & 74.5 & 83.8 & 2.0 & 13.0 & 45.3 & 74.1 & 83.7 & 2.0 & 9.2  \\
OmniVL\cite{Wang2022OmniVLOF} & 47.8 & 74.2 & 83.8 & - & - & - &- &- &- &- \\
HBI\cite{jin2023video} & 48.6 & 74.6 & 83.4 & 2.0 & 12.0 & 46.8 &74.3 &\textbf{84.3} &2.0 &8.9 \\
\hline
Ours(Base) & \textbf{49.6} & \textbf{75.6} & \textbf{84.9} & \textbf{2.0} & \textbf{12.0} & \textbf{47.0} & \textbf{76.7} & 84.0 & \textbf{2.0} & \textbf{8.8}  \\
Ours(Large)& \textbf{54.0} & \textbf{77.5} & \textbf{85.3} & \textbf{1.0} & \textbf{11.8} & \textbf{51.3} & \textbf{78.3} & \textbf{85.4} & \textbf{1.0} & \textbf{8.5}  \\ 
\hline

\end{tabular}
}
\vspace{1mm}
\caption{ Retrieval performance comparison on MSR-VTT. 
Here we report results without any post-processing operations (e.g., DSL ~\cite{cheng2021improving} or QB-Norm ~\cite{bogolin2021cross}) during inference. 
}
\vspace{-5mm}
\label{table:msrvtt}
\end{table}

\textbf{Evaluation Metrics.} To evaluate the performance of our model, we use standard retrieval metrics: recall at rank K (R@K, higher is better), median rank (MdR, lower is better), and mean rank (MnR, lower is better). R@K calculates the percentage of test samples that find correct results within the top K query samples. We select K=1,5,10 in the experiment. Mdr and MnR represent the median and mean of the ground-truth results in the retrieval ranking, respectively.

\textbf{Implementation Details.} 
In the first stage, we use TS2Net (ViT-B/32)~\cite{liu2022ts2} as the network for cosine similarity computation, and the weights of the visual encoder and text encoder are initialized from the publicly available CLIP checkpoints. 
In the second stage, the visual encoder for cross attention can be ViT-B or ViT-G, both are also initialized from CLIP. And the weights of spatial and temporal cross attention are initialized from BLIP-2. 
The other modules are initialized randomly. We set M=4 in token selector and K=15 in re-ranking stage. The max query text length and max video frame length are 32 and 12 for MSR-VTT, MSVD, VATEX, and LSMDC. For DiDeMo, the max query text length and max video frame length are set 64 and 64. We train the model with the Adam ~\cite{kingma2014adam}. The initial learning rate of the trainable modules is 1e-4. Because the vision encoder of CrossTVR is frozen, we can precompute and cache the vision tokens in memory for efficient training. All learning rates adopt the cosine schedule with a warm up setup. We use one 80G NVIDIA A100 card for training and the batch size is set to 128 for all datasets. 
\subsection{Comparison with State-of-the-arts}\vspace{-2mm}
\textbf{MSR-VTT.} Table \ref{table:msrvtt} compares our method to existing methods on the MSRVTT dataset. Our method is superior across all the metrics. Especially reaching 49.6 on Text2Video (T2V) R@1 and 47.0 on Video2Text (V2T) R@1 using the ViT-B backbone, surpassing the existing methods by 1.0\% on T2V R@1. With a large backbone ViT-G, the performance is further improved by 4.4\% on T2V R@1.

\textbf{Other benchmarks.} Our method outperforms existing methods on other benchmarks, as shown in Table~\ref{Table: msvd} to Table~\ref{Table: VATEX}. Our T2V R@1 on the four datasets reach 49.7 (+2.5\%) on MSVD, 26.8 (+1.6\%) on LSMDC, 47.1 (+1.9\%) on DiDeMo, and 64.1 (+5.0\%) on VATEX, respectively. The consistent improvement in different datasets demonstrates the generalization and robustness of our method. 
Particularly, the performance improvement of our method is more obvious in the VATEX dataset where the text description is more semantically complex and diverse. We consider this because our fine-grained CrossTVR enhances the understanding of complex statements.

\begin{table}[t]
	\centering
	\begin{minipage}{0.45\textwidth}
		\centering
            \renewcommand{\arraystretch}{1.1}
		\makeatletter\def\@captype{table}\makeatother
            \resizebox{1.1\textwidth}{!}{
             \setlength{\tabcolsep}{0.5mm}{
            \begin{tabular}{c|ccc|ccc}
            \hline
            \multirow{2}{*}{\begin{tabular}[c]{@{}c@{}}Method \end{tabular}} & \multicolumn{3}{c|}{Text2Video} & \multicolumn{3}{c}{Video2Text} \\ \cline{2-7} 
                                                                                   & R@1       & R@5       & MnR     & R@1       & R@5      & MnR     \\ \hline
            CLIP\cite{radford2021learning}                                         & 37.0      & 64.1      & -       & -         & -        & -       \\
            Frozen\cite{bain2021frozen}                                            & 33.7      & 64.7      & -       & -         & -        & -       \\
            CLIP4Clip\cite{luo2021clip4clip}                                       & 45.9      & 74.9      & 10.4    & 51.0      & 76.3     & 9.1     \\
            CenterCLIP\cite{zhao2022centerclip}                                    & 47.6      & 77.5      & 9.8     & 54.2      & 78.4     & 7.6     \\
            CAMoE\cite{cheng2021improving}                                         & 46.9      & 76.1      & 9.8     & -         & -        & -       \\
            CLIP2Video\cite{fang2021clip2video}                                    & 47.0      & 76.8      & 9.6     & 58.7      & 85.6     & 4.3     \\
            X-CLIP\cite{ma2022x}                                                    & 47.1      & 77.8     & 9.5     & 60.9      & 87.8     & 4.7      \\ 
            X-Pool\cite{gorti2022x}                                                & 47.2      & 77.4      & 9.3     & -      & -     & -     \\\hline
            Ours(Base)                            & \textbf{49.7}       & \textbf{79.0}     & \textbf{9.1}     & \textbf{68.7}      &  \textbf{89.6}   & \textbf{3.0}     \\
            Ours(Large)                            &  \textbf{55.0}     &   \textbf{81.9}    & \textbf{9.1}     & \textbf{74.3}      &  \textbf{95.5}    & \textbf{2.0}   \\ 
            \hline
            \end{tabular}
            }}

            \vspace{1mm}
            \caption{Retrieval performance on MSVD} 
            \label{Table: msvd}
	\end{minipage}\qquad
 	\begin{minipage}{0.45\textwidth}
		\centering
              \renewcommand{\arraystretch}{1.1}

		\makeatletter\def\@captype{table}\makeatother
            \resizebox{1.1\textwidth}{!}{
            \setlength{\tabcolsep}{0.5mm}{
            \begin{tabular}{c|ccc|ccc}
            \hline
            \multirow{2}{*}{\begin{tabular}[c]{@{}c@{}}Method \end{tabular}} & \multicolumn{3}{c|}{Text2Video} & \multicolumn{3}{c}{Video2Text} \\ \cline{2-7} 
                                                                                    & R@1       & R@5      & MnR      & R@1      & R@5      & MnR      \\ \hline
            CLIP\cite{radford2021learning}                               & 11.3      & 22.7     & -         & -        & -        & -        \\
            Frozen\cite{bain2021frozen}                                  & 15.0      & 30.8     & -         & -        & -        & -        \\
            CLIP4Clip\cite{luo2021clip4clip}                             & 22.6      & 41.0     & 61.0      & 20.8     & 39.0     & 54.2     \\
            CenterCLIP\cite{zhao2022centerclip}                          & 21.9      & 41.1     & 55.6     & 21.1     & 41.2     & 48.7     \\
            CAMoE\cite{cheng2021improving}                               & 22.5      & 42.6     & 56.5        & -        & -        & -      \\
            X-CLIP\cite{ma2022x}                                          & 23.3      & 43.0     & 56.0       & 22.5     & 42.2     & 50.7   \\
            TS2-Net\cite{liu2022ts2}                                    & 23.4      & 42.3     & 56.9      & -        & -        & -        \\ 
            X-Pool\cite{gorti2022x}                                     & 25.2      & 43.7     & 53.2      & -     & -     & -     \\\hline
            Ours(Base)                    & \textbf{26.8 }       & \textbf{45.3}     &  \textbf{55.2}     &   \textbf{23.8}    &   \textbf{45.0}   &   \textbf{46.6}   \\
            Ours(Large)                  & \textbf{27.7}       &  \textbf{48.5 }   &  \textbf{52.8 }   &   \textbf{29.7}    & \textbf{47.4 }     &   \textbf{46.2}  \\ 
            \hline
            \end{tabular}
            }}
            \vspace{1mm}
            \caption{Retrieval performance on LSMDC} 
            \label{Table: lsmdc}

	\end{minipage}
        \vspace{-1mm}
        
		\begin{minipage}{0.45\textwidth}
		\centering
            \renewcommand{\arraystretch}{1.2}
		\makeatletter\def\@captype{table}\makeatother
            \resizebox{1.1\textwidth}{!}{
             \setlength{\tabcolsep}{0.5mm}{
             \begin{tabular}{c|ccc|ccc}
            \hline
            \multirow{2}{*}{\begin{tabular}[c]{@{}c@{}}Method \end{tabular}} & \multicolumn{3}{c|}{Text2Video} & \multicolumn{3}{c}{Video2Text} \\ \cline{2-7} 
                                                                                     & R@1       & R@5       & MnR     & R@1       & R@5      & MnR     \\ \hline
            ClipBERT\cite{lei2021less}                                                & 20.4      & 44.5      & -      & -         & -        & -       \\
            Frozen\cite{bain2021frozen}                                               & 34.6      & 65.0      & -       & -         & -        & -       \\
            CLIP4Clip\cite{luo2021clip4clip}                                          & 43.4      & 70.2      & 17.5     & 42.5      & 70.6     & 11.6     \\
            CAMoE\cite{cheng2021improving}                                            & 43.8      & 71.4      & 16.3     & 45.5      & -        & \textbf{10.2}     \\
            X-CLIP\cite{ma2022x}                                                      & 45.2      & 74.0      & \textbf{14.6}       & 43.1      & 72.2     & 10.9    \\
            TS2-Net\cite{liu2022ts2}                                                 & 41.8    & 71.6      & 14.8     & -         & -        & -       \\ \hline
            Ours(Base)                   &  \textbf{47.1}     & \textbf{73.9}    & 17.9       & \textbf{45.4}      & \textbf{72.6}     &  12.1   \\
            Ours(Large)                &  \textbf{55.0 }    & \textbf{77.6}    &  17.6     &   \textbf{51.3}      & \textbf{76.7 }    &  11.6 \\ 
            \hline
            \end{tabular}
            }}
            \vspace{1mm}
            \caption{Retrieval performance on DiDeMo} 
            \label{Table: DiDeMo}

	\end{minipage}\qquad
 	\begin{minipage}{0.45\textwidth}
		\centering
              \renewcommand{\arraystretch}{1.2}

		\makeatletter\def\@captype{table}\makeatother
            \resizebox{1.1\textwidth}{!}{
            \setlength{\tabcolsep}{0.5mm}{
            \begin{tabular}{c|ccc|ccc}
            \hline
            \multirow{2}{*}{\begin{tabular}[c]{@{}c@{}}Method\end{tabular}} & \multicolumn{3}{c|}{Text2Video} & \multicolumn{3}{c}{Video2Text} \\ \cline{2-7} 
                                                                                    & R@1       & R@5       & MnR     & R@1       & R@5      & MnR     \\ \hline
            HGR\cite{chen2020fine}                                                  & 35.1      & 73.5      & -        & -         & -        & -       \\
            CLIP\cite{radford2021learning}                                          & 39.7      & 72.3      & 12.8     & 52.7      & 88.8     & 3.8     \\
            SUPPORT\cite{patrick2020support}                                        & 44.9      & 82.1      & 3.9      & 58.4      & 84.4     & -     \\
            CLIP4Clip\cite{luo2021clip4clip}                                        & 55.9      & 89.2      & 3.9      & 73.2      & 97.1     & 1.7     \\
            CLIP2Video\cite{fang2021clip2video}                                     & 57.3      & 90.0      & 3.6     & 76.0      & 97.7     & 1.5     \\
            TS2-Net\cite{liu2022ts2}                      & 59.1      & 90.0      & 3.5     & -      & -     & -     \\ \hline
            Ours(Base)            & \textbf{64.1}      & \textbf{91.5}      & \textbf{3.3}     & \textbf{79.7 }     & \textbf{97.2}     & \textbf{1.6}     \\
            Ours(Large)         & \textbf{71.1}      & \textbf{94.0 }     & \textbf{3.0}     & \textbf{86.4}      & \textbf{99.2}     & \textbf{1.3 }    \\ 
            \hline
            \end{tabular}
            }}
            \vspace{1mm}
            \caption{Retrieval performance on VATEX} 
            \label{Table: VATEX}

	\end{minipage}
\vspace{-8mm}
\end{table}

\begin{figure}[h]
  \centering
  \vspace{-5pt}
  \includegraphics[width=0.9\textwidth]{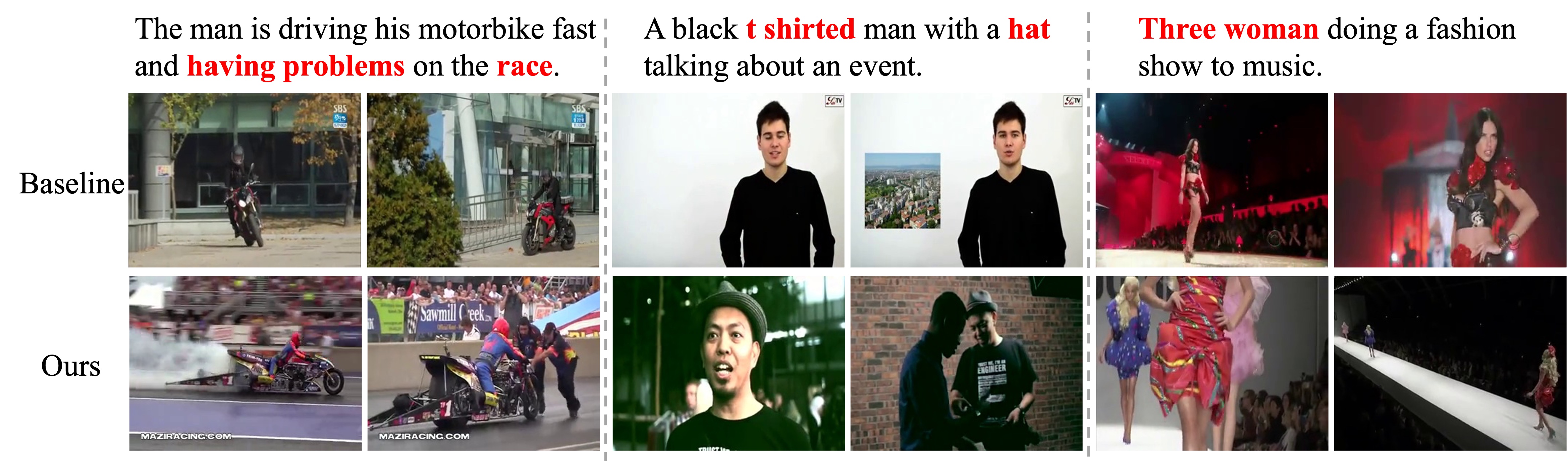}
  \caption{The visualization of text video retrieval is in MSRVTT. The top is the retrieved text, and the last two rows are the rank1 retrieval results of baseline and our method respectively. We highlight the words in the text that do not match the retrieval in red. We can see that our method retrieves fine-grained vision features based on the query text.}\label{fig: visualize}
  \vspace{-6mm}
\end{figure}

\subsection{Qualitative Analysis}\vspace{-2mm}\label{sec:qualitative} 
To qualitatively measure the effectiveness of our proposed CrossTVR, we visualize the text-to-video results  for several examples of the MSRVTT dataset in Figure \ref{fig: visualize}. These results demonstrate the ability of our CrossTVR to more accurately understand the text and video content. Specifically, in the first example, baseline only captures `the man is driving his motorbike', while our CrossTVR further understands `having problems on the race'. In the second example, our CrossTVR correctly retrieves the small object `hat' and can distinguish `shirt' and `t-shirt'. In the third example, on the hierarchical semantics `three women', which requires cognition of textual videos, our model also correctly retrieves. Compared with the cosine similarity-based baseline, our method is more capable of understanding the video text content and uncovering detailed information.


\subsection{Ablation Study}\vspace{-2mm}\label{sec:ablation}

\paragraph{Spatial and temporal text attention module}
We analyze the impact of temporal and spatial text attention on our CrossTVR Large model, and the results are presented in Table~\ref{table spatial temporal}. We denote the first stage TS2Net as the baseline method.
When we add temporal text attention and spatial text attention separately, we observe improvements in T2V R@1 by 4.3\% and 4.5\%, respectively. 
By utilizing both modules, CrossTVR achieves the highest retrieval accuracy with a significant improvement in T2V R@1 by 7\%. 
The statistics demonstrate similar performance for the V2T task as well. 
These findings highlight the effectiveness of our spatial temporal modules for T2V and V2T retrieval tasks.

\vspace{-2mm}
\begin{table}[h]\small
\centering
\renewcommand{\arraystretch}{1.2}

\begin{tabular}{ccc|ccc|ccc}
\hline
\multirow{2}{*}{Baseline} & \multirow{2}{*}{\begin{tabular}[c]{@{}c@{}}Spatial\end{tabular}} & \multirow{2}{*}{\begin{tabular}[c]{@{}c@{}}Temporal\end{tabular}} & \multicolumn{3}{c|}{Text2Video} & \multicolumn{3}{c}{Video2Text} \\ \cline{4-9} 
            &              &                    & R@1       & R@5      & MnR      & R@1       & R@5      & MnR     \\ \hline
\checkmark &              &                     & 47.0      & 74.5     & 13.0     & 45.3      & 74.1     & 9.2     \\
\checkmark & \checkmark   &                     & 51.3      & 75.9     & 12.0     & 50.3      & 76.9     & 8.6    \\
\checkmark &              & \checkmark          & 51.5      & 76.8     & 11.9     & 51.0      & 74.8     & 8.7     \\
\checkmark & \checkmark   & \checkmark          & 54.0      & 77.5     & 11.8     & 51.3      & 78.3     & 8.5     \\ \hline
\end{tabular}
\vspace{1mm}
\caption{ Ablation study of spatial and temporal text attention module of our method on MSRVTT.}
\label{table spatial temporal}
\vspace{-7mm}
\end{table}

\paragraph{Vision encoder scalability} 
Table~\ref{table train cost} illustrates the GPU memory cost and training speed of end-to-end finetuning method and CrossTVR with different vision encoders. We re-implement the representative similarity based approach CLIP4Clip~\cite{luo2021clip4clip} with ViT-G and train it on 8 80G NVIDIA A100 cards with the maximum batch size before run out of GPU memory. From ViT-B to ViT-G, CLIP4Clip increases more than 10$\times$ GPU memory while our method only increases 22\% memory. With the same ViT-G, we can see that our method reduces 91\% memory consumption and 58\% training time, which shows our method scales better to large models than finetuning methods.
\vspace{-4mm}
\begin{table}[h]\small
      \centering
      \renewcommand{\arraystretch}{1.2}

        \begin{tabular}{c|c|c|c|c|c}
        \hline
        Method  & 
        \begin{tabular}[c]{@{}c@{}}Vision \\ Encoder \end{tabular} & 
        \begin{tabular}[c]{@{}c@{}}Text2Video \\ R@1 \end{tabular} &
        \begin{tabular}[c]{@{}c@{}}Batch \\  Size\end{tabular} & \begin{tabular}[c]{@{}c@{}}Memory \\ Cost (GB)\end{tabular} & 
        \begin{tabular}[c]{@{}c@{}}Training Cost \\(GPU hour)\end{tabular}\\ \hline
        CLIP4Clip~\cite{luo2021clip4clip} & ViT-B/32 & 44.5 &  128  & 41 & 9 \\
        CLIP4Clip~\cite{luo2021clip4clip} & ViT-G/14 & 47.4 &  48  & 546 & 50\\
        Ours & ViT-B/32 & 49.6 & 128  & 41  & 18 \\
        Ours & ViT-G/14 & 54.0 & 128  & 50  & 21 \\ \hline
        \end{tabular}
        \vspace{1mm}
        \captionof{table}{From ViT-B to ViT-G, the increase of memory cost and training cost of our method is smaller than end-to-end finetuned CLIP4Clip. We report the accuracy on MSRVTT dataset. 
        } 
        \vspace{-6mm}
        \label{table train cost}
\end{table}

\paragraph{Collaborating with different cosine similarity based methods} 
Our CrossTVR seamlessly integrates with various cosine similarity-based methods~\cite{luo2021clip4clip,liu2022ts2,gorti2022x}, thanks to its decoupled video text cross attention header. 
To demonstrate the generalization and robustness of our cross attention module, we evaluate CrossTVR Base with three state-of-the-art cosine similarity-based methods ~\cite{luo2021clip4clip,liu2022ts2,gorti2022x}, as shown in Table~\ref{table different network}.
The results reflect obvious improvements of 2.5\% for CLIP4Clip~\cite{luo2021clip4clip}, 2.6\% for TS2Net~\cite{liu2022ts2}, and 1.2\% for X-pool~\cite{gorti2022x} on T2V R@1 when combined with our cross attention module.
These notable gains across the three methods highlight the wide applicability and flexibility of our cross attention module. 
Furthermore, there is almost no difference in inference time with and without our cross attention module. 
We firmly believe that our cross attention module has the potential to act as a powerful performance-boosting module for most cosine similarity-based TVR methods, present and future.

\paragraph{Sharing parameters in cross attention} 
In our CrossTVR approach, the spatial text attention and temporal text attention modules share all parameters. 
We conduct an ablation study, depicted in Table~\ref{table share parameters spatial temporal}, to investigate whether parameter sharing can improve retrieval performance beyond its parameter efficiency advantage. 
The results indicate that sharing parameters lead to a higher retrieval performance compared to unshared parameters, with an improvement of 2.6\% T2V R@1 compared to the baseline. 
In contrast, unshared parameters only improved retrieval performance by 1.6\% T2V R@1. 
These findings suggest that the spatial and temporal cross attention modules can benefit from each other when parameters are shared.

\begin{table}[!h]\small
\vspace{-3mm}
\centering
\renewcommand{\arraystretch}{1.2}
\begin{tabular}{c|ccc|ccc|c}
\hline
\multirow{2}{*}{Method} & \multicolumn{3}{c|}{Text2Video} & \multicolumn{3}{c|}{Video2Text} & \multirow{2}{*}{Time (s)} \\ \cline{2-7} 
                        & R@1       & R@5      & MnR      & R@1      & R@5      & MnR                      \\ \hline
CLIP4Clip               & 44.5      & 71.4     & 15.3     & 42.7     & 70.9     & 11.6  &  7.06 \\
CLIP4Clip+CrossTVR      & 47.0      & 73.8     & 15.5     & 44.1     & 71.9     & 11.9  &  7.21 \\ \hline
TS2Net                  & 47.0      & 74.5     & 13.0     & 45.3     & 74.1     & 9.2   &  8.24 \\
TS2Net+CrossTVR         & 49.6      & 75.5     & 12.1     & 47.0     & 76.1     & 8.8   &  8.39 \\ \hline
Xpool                   & 46.9      & 72.8     & 14.3     & 44.4     & 73.3     & 9.0   &  12.17 \\
Xpool+CrossTVR          & 48.1      & 74.5     & 14.1     & 47.8     & 75.2     & 8.8   &  12.38 \\ \hline
\end{tabular}
\vspace{1mm}
\caption{ Ablation study about different cosine similarity networks on MSRVTT dataset. Time shows inference speed for indexing 1000 videos with one query text on one 80G NVIDIA A100 card.} 
\label{table different network}
\vspace{-7mm}
\end{table}




\paragraph{Inference speed}
The last column of Table~\ref{table different network} presents the time it takes to encode and retrieve 1000 videos with one text on the MSRVTT dataset using ViT-B. 
The methods CLIP4Clip~\cite{luo2021clip4clip}, TS2-Net~\cite{liu2022ts2}, and X-pool~\cite{gorti2022x} take 7.06s, 8.24s, and 12.17s, respectively. 
In comparison, our CrossTVR method requires an additional time of 0.15s, 0.15s, and 0.21s, respectively. 
Despite this additional computational cost of about $2\%$, our approach achieves an obvious improvement of retrieval accuracy.

\vspace{-5mm}
\begin{figure}[ht]
	\begin{minipage}{0.37\textwidth}
        \centering
      \includegraphics[width=0.95\textwidth]{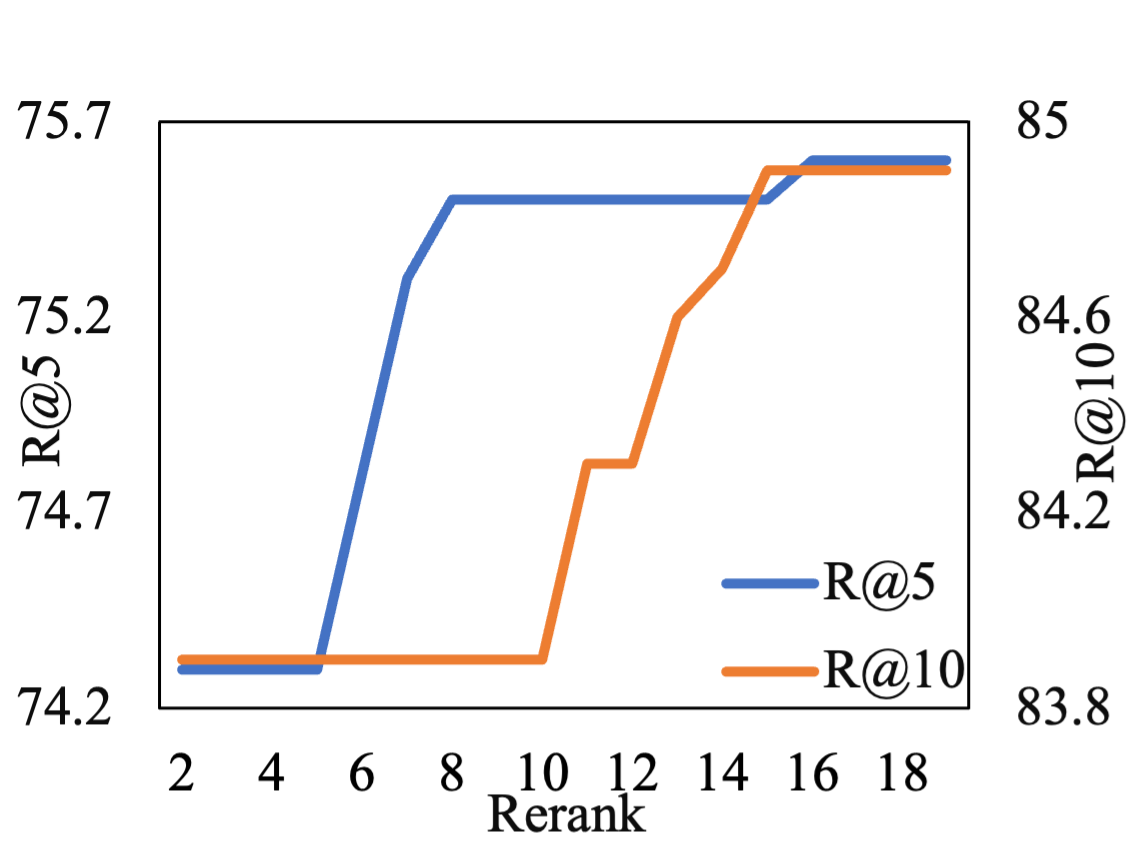}
      \vspace{-3mm}
      \caption{Different re-rank numbers.}\label{fig: abation rerank}
    \end{minipage}\quad	    
    \begin{minipage}{0.6\textwidth}
\centering
\renewcommand{\arraystretch}{1.2}
\resizebox{0.95\textwidth}{!}{
\begin{tabular}{c|ccc|ccc}
\hline
\multirow{2}{*}{Method}       & \multicolumn{3}{c|}{Text2Video} & \multicolumn{3}{c}{Video2Text} \\ \cline{2-7} 
                              & R@1       & R@5      & MnR      & R@1      & R@5      & MnR      \\ \hline
Baseline                      & 47.0      & 74.5     & 13.0     & 45.3     & 74.1     & 9.2      \\
Unsharing & 48.6      & 74.9     & 12.2     & 46.3     & 75.0     & 8.8      \\
Sharing  & 49.6      & 75.5     & 12.1     & 47.0     & 76.1     & 8.8      \\ \hline
\end{tabular}}

\vspace{0.8mm}
\captionof{table}{Ablation study about whether sharing parameters between spatial and temporal cross attention on MSRVTT dataset.}
\label{table share parameters spatial temporal}
    \end{minipage}    

\vspace{-5mm}
\end{figure}

\paragraph{Benefits of re-ranking}
The re-rank number $K$ is a crucial factor in determining retrieval performance. 
To investigate this relationship, we perform experiments on ViT-B network using the MSRVTT dataset, varying the re-rank number from 2 to 19. 
As depicted in Figure~\ref{fig: abation rerank}, we observe a gradual increase in R@5 and R@10 T2V scores as $K$ increases, indicating that our method is effective across various ranks. 
However, excessively large values of $K$ can decrease retrieval efficiency. 
To strike a balance between efficiency and accuracy, we set $K$=15 for all our experiments.


%% file: conclusion.tex
\section{Conclusion and Discussion of Broader Impact}\vspace{-2mm}
In this paper, we present CrossTVR to explore fine-grained text-video interaction information with decoupled video text cross attention transformer module. 
Our CrossTVR method achieves superior performances on various TVR benchmarks and scales well to large pre-trained vision models. However, since current SOTA cosine similarity methods require end-to-end fine-tuning, we can not freeze the first stage's vision encoder. In the future, we expect an efficient unified framework which can freeze the vision encoder end to end. CrossTVR is an effective method for text video retrieval, which could increase the risk of video understanding model or its outputs being used incorrectly, such as for unauthorized surveillance.

%% file: appendix.tex
\section*{Appendix}
\subsection*{Sharing vision encoder} In CrossTVR Base, we use one ViT-B/32 for the fist stage cosine similarity and another ViT-B/32 for the second stage cross attention. We can also share a single vision encoder ViT-B/32 in both stages to further increase the retrieval speed. In Table\ref{table share vision encoder}, we conduct experiments on the parameter sharing of visual encoders between the two headers. 
When sharing the visual encoder, the T2V R@1 is the same as that when unsharing, both being 49.6. Additionally, the differences in other metrics also are not significant. And the speed of inference is reduced by 0.03s for shared encoder compared to unshared.
Based on this finding, we can choose the vision encoder sharing setting to reduce the model size.

\begin{table}[!h]\small
\centering
\renewcommand{\arraystretch}{1.2}
\resizebox{0.8\textwidth}{!}{
\begin{tabular}{c|ccc|ccc|c}
\hline
\multirow{2}{*}{Method}       & \multicolumn{3}{c|}{Text2Video} & \multicolumn{3}{c|}{Video2Text} & \multirow{2}{*}{Time (s)}  \\ \cline{2-7} 
                              & R@1       & R@5      & MnR      & R@1      & R@5      & MnR      \\ \hline
Baseline                      & 47.0      & 74.3     & 13.0     & 45.3     & 74.1     & 9.2  &   8.24  \\
Unsharing vision encoder    & 49.6      & 75.5     & 12.1       & 47.0     & 76.1     & 8.8  &   8.39\\
Sharing vision encoder       & 49.6     & 75.6      & 12.0     & 47.0      & 76.7     & 8.8  &  8.36\\ \hline
\end{tabular}}

\vspace{0.8mm}
\caption{ Ablation study about sharing vision encoders on MSR-VTT dataset.Time shows inference speed for indexing 1000 videos with one query text on one 80G NVIDIA A100 card.}
\label{table share vision encoder}
\vspace{-15pt}
\end{table}
\subsection*{DSL Pos-processing} The hub phenomenon often occurs in retrieval tasks, where data appears in the k-nearest neighbors of other data. To solve this problem, \cite{cheng2021improving} proposed DSL, using the intrinsic prior of each pair in a batch, and modified the similarity matrix with inverted softmax to achieve the optimal matching of pairs.  We also implement DSL in our inference, and the results are shown in Table \ref{Table: msrvtt dsl} to \ref{Table: VATEX dsl}. Our results achieve further improvement on each dataset.

\begin{table}[h]
	\centering
	\begin{minipage}{0.45\textwidth}
		\centering
            \renewcommand{\arraystretch}{1.1}
		\makeatletter\def\@captype{table}\makeatother
            \resizebox{1.1\textwidth}{!}{
             \setlength{\tabcolsep}{0.5mm}{
           \begin{tabular}{c|ccc|ccc}
            \hline
            \multirow{2}{*}{\begin{tabular}[c]{@{}c@{}}Method \end{tabular}} & \multicolumn{3}{c|}{Text2Video} & \multicolumn{3}{c}{Video2Text} \\ \cline{2-7} 
                                                                                    & R@1       & R@5      & MnR      & R@1      & R@5      & MnR      \\ \hline
              CrossTVR(Base)                    &  49.6        &  75.6     &    12.0    &     47.0   &     76.7   &     8.8   \\
              CrossTVR$^*$((Base)            &   52.8        &   78.9     &    11.2     &     52.5   &     77.3  &     8.9   \\ \hline
              CrossTVR(Large)                  &   54.0       &    77.5   &    11.8    &    51.3    &   78.3      &    8.5  \\ 
              CrossTVR$^*$(Large)                 & 57.0      &  81.0    &  12.2    &   56.6    & 79.0      &   10.1 \\
            \hline
            \end{tabular}
            }}
            \vspace{1mm}
            \caption{Retrieval performance using DSL post-processing on MSR-VTT. * denotes the use of DSL.} 
            \label{Table: msrvtt dsl}
	\end{minipage}\qquad
 	\begin{minipage}{0.45\textwidth}
		\centering
              \renewcommand{\arraystretch}{1.1}

		\makeatletter\def\@captype{table}\makeatother
            \resizebox{1.1\textwidth}{!}{
            \setlength{\tabcolsep}{0.5mm}{
             \begin{tabular}{c|ccc|ccc}
            \hline
            \multirow{2}{*}{\begin{tabular}[c]{@{}c@{}}Method \end{tabular}} & \multicolumn{3}{c|}{Text2Video} & \multicolumn{3}{c}{Video2Text} \\ \cline{2-7} 
                                                                                   & R@1       & R@5       & MnR     & R@1       & R@5      & MnR     \\ \hline
              CrossTVR(Base)          &   49.7      &   79.0     &   9.1     &   68.7      &    89.6   &   3.0     \\
              CrossTVR$^*$(Base)          &   52.7      &   80.8     &   9.1     &   76.7      &   93.6   &   2.2     \\ \hline
              CrossTVR(Large)             &    55.0     &     81.9    &   9.1     &   74.3      &    95.5    &   2.0   \\ 
              CrossTVR$^*$(Large)       &  57.9     &   83.1    & 10.1     & 81.5      &  96.1    & 1.8 \\
            \hline
            \end{tabular}
            }}

            \vspace{1mm}
            \caption{Retrieval performance using DSL post-processing on MSVD. * denotes the use of DSL.} 
            \label{Table: msvd dsl}

	\end{minipage}
        \vspace{-1mm}
        
		\begin{minipage}{0.45\textwidth}
		\centering
            \renewcommand{\arraystretch}{1.2}
		\makeatletter\def\@captype{table}\makeatother
            \resizebox{1.1\textwidth}{!}{
             \setlength{\tabcolsep}{0.5mm}{
             \begin{tabular}{c|ccc|ccc}
            \hline
            \multirow{2}{*}{\begin{tabular}[c]{@{}c@{}}Method \end{tabular}} & \multicolumn{3}{c|}{Text2Video} & \multicolumn{3}{c}{Video2Text} \\ \cline{2-7} 
                                                                                     & R@1       & R@5       & MnR     & R@1       & R@5      & MnR     \\ \hline
              CrossTVR(Base)     &    47.1     &   73.9    & 17.9       &   45.4      &   72.6     &  12.1   \\
              CrossTVR(Base)      &    51.2     &   75.1    & 15.9       &   51.9      &   76.7     &  11.2   \\ \hline
              CrossTVR(Large)        &    55.0    &   77.6    &  17.6     &     51.3      &   76.7     &  11.6 \\ 
              CrossTVR$^*$(Large)    &  57.5     & 77.8    &  19.1     &   56.7      & 77.3     &  14.7  \\  
            \hline
            \end{tabular}
            }}
            \vspace{1mm}
            \caption{Retrieval performance using DSL post-processing on DiDeMo. * denotes the use of DSL.} 
            \label{Table: DiDeMo dsl}

	\end{minipage}\qquad
 	\begin{minipage}{0.45\textwidth}
		\centering
              \renewcommand{\arraystretch}{1.2}

		\makeatletter\def\@captype{table}\makeatother
            \resizebox{1.1\textwidth}{!}{
            \setlength{\tabcolsep}{0.5mm}{
            \begin{tabular}{c|ccc|ccc}
            \hline
            \multirow{2}{*}{\begin{tabular}[c]{@{}c@{}}Method\end{tabular}} & \multicolumn{3}{c|}{Text2Video} & \multicolumn{3}{c}{Video2Text} \\ \cline{2-7} 
                                                                                    & R@1       & R@5       & MnR     & R@1       & R@5      & MnR     \\ \hline
              CrossTVR(Base)            &   64.1      &   91.5      &   3.3     &   79.7      &   97.2     &   1.6     \\
              CrossTVR$^*$(Base)        &   66.2      &   92.2      &   3.3     &   87.4      &   98.1     &   1.4     \\ \hline
              CrossTVR(Large)         &   71.1     &   94.0      &   3.0     &   86.4      &   99.2     &   1.3     \\ 
              CrossTVR$^*$(Large)                             & 73.7      & 94.3      & 2.9     & 90.9      & 99.2     & 1.3\\
            \hline
            \end{tabular}
            }}
            \vspace{1mm}
            \caption{Retrieval performance using DSL post-processing on VATEX. * denotes the use of DSL.} 
            \label{Table: VATEX dsl}

	\end{minipage}
\vspace{-8mm}
\end{table}

\begin{figure}[t]
  \centering
  \vspace{-5pt}
  \includegraphics[width=0.98\textwidth]{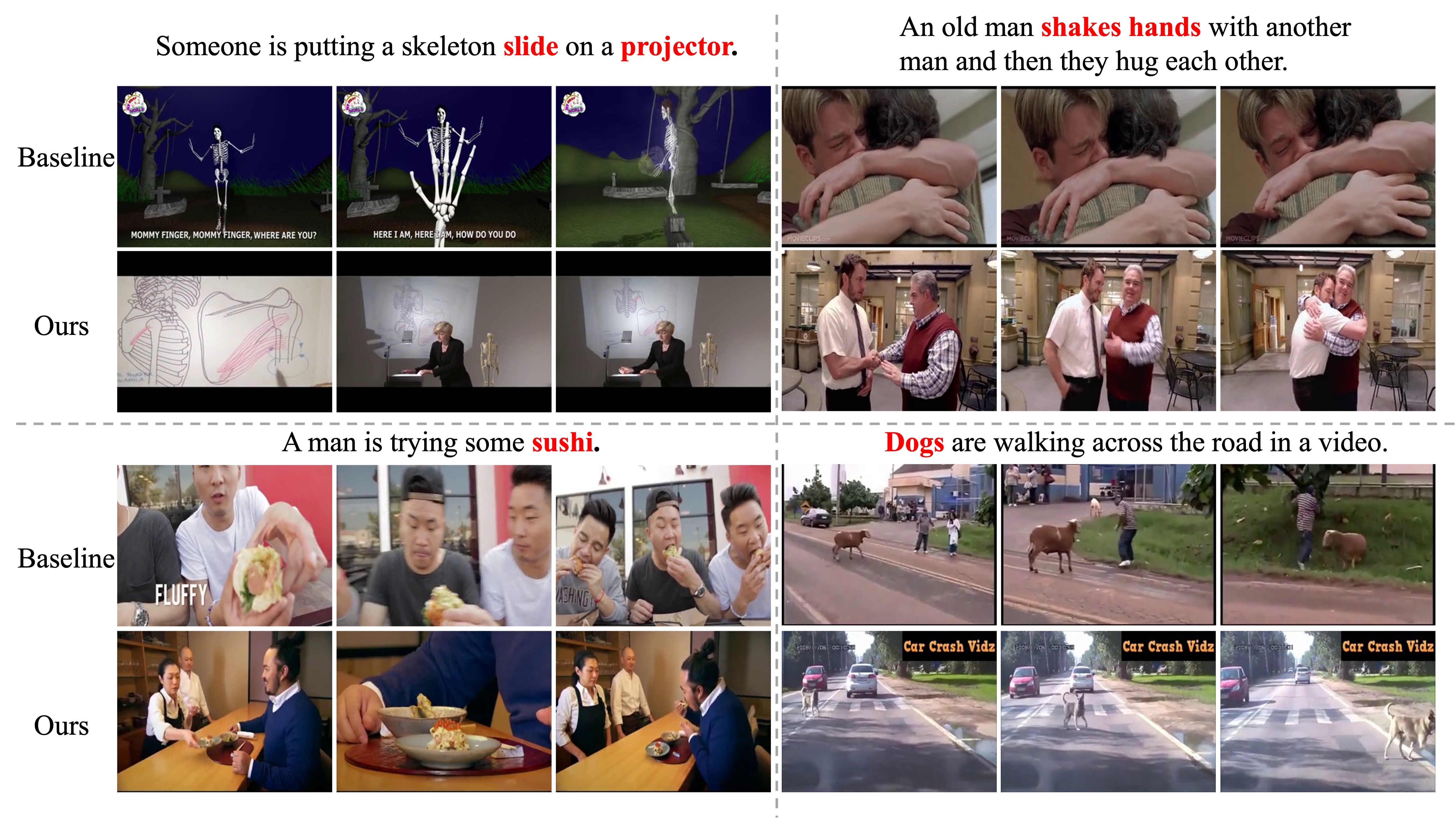}
  \caption{ Text2Video retrieval examples on MSR-VTT. The text is the retrieved text, and the following two lines are the rank1 retrieval results of baseline and our method respectively. We highlight the words in the text that do not match the retrieval in red.}\label{fig: visualize t2v}
  \vspace{1mm}
\end{figure}

\begin{figure}[t]
  \centering
  \vspace{-5pt}
  \includegraphics[width=0.98\textwidth]{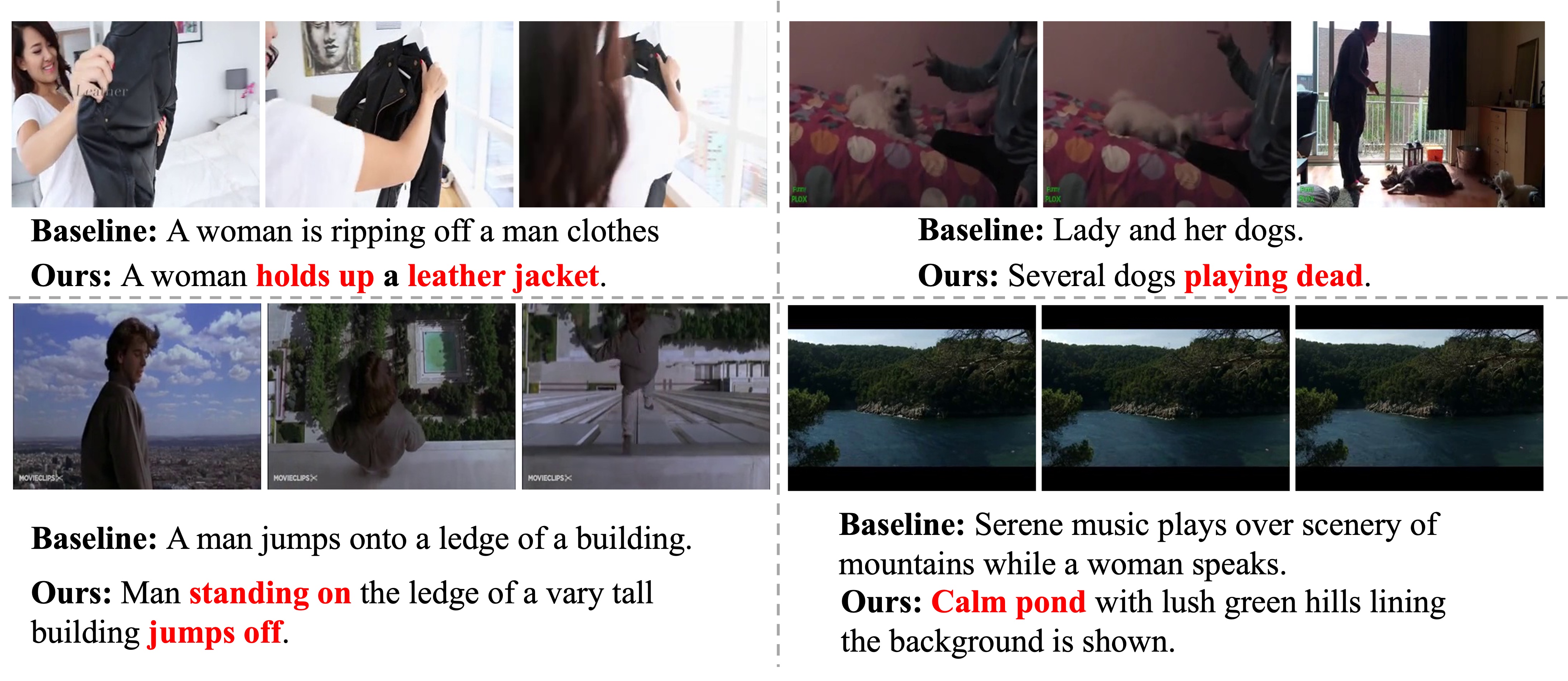}
  \caption{ Video2Text retrieval examples on MSR-VTT.  The video is the retrieved video, and the following two lines are the rank1 retrieval results of baseline and our method respectively. We highlight the keywords that match the video in red.}\label{fig: visualize v2t}
  \vspace{1mm}
\end{figure}

\subsection*{More Qualitative Results} We show more qualitative results of Text2Video Figure \ref{fig: visualize t2v}. In the top left example, the baseline only recognizes the `skeleton' and ignores the action `slid on a projector'. In the top right example, our method is able to capture the action `shakes hands'. These two examples illustrate the advantages of our method in capturing subtle movements. In the bottom left and bottom right examples, our method is able to identify small objects, such as `sushi' and `hotdog', `dog' and `goat'.

Further, we present the qualitative results of the Video2Text retrieval in Figure \ref{fig: visualize v2t}. In the top left example our method recognizes the more detailed feature `hold up a leather jacked' while baseline only recognizes `clothes'. Both the top right and bottom left examples demonstrate the strength of our method in identifying subtle movement, successfully identifying `playing dead' and `jumps off'. In the bottom right example, the baseline only identifies the `mountain' while our method also identifies the `calm pond'.